# Multimodal Survival Modeling and Fairness-Aware Clinical Machine Learning for 5-Year Breast Cancer Risk Prediction

Author: Toktam Khatibi, Associate Professor, Faculty of Industrial and Systems Engineering, Tarbiat Modares University, Tehran, Iran, email: toktam.khatibi@modares.ac.ir; toktamk@gmail.com

## Abstract

Clinical risk prediction models often underperform in real-world settings due to poor calibration, limited transportability, and subgroup-level performance disparities. These challenges are amplified in high-dimensional multimodal cancer datasets characterized by complex feature interactions and a p≫n structure. We present a fully reproducible multimodal clinical machine learning framework for 5-year overall survival prediction in breast cancer, integrating clinical variables with high-dimensional transcriptomic and copy-number alteration (CNA) features using the METABRIC cohort.

Following variance- and sparsity-based filtering and truncated singular value decomposition, models were trained using stratified train/validation/test splits with hyperparameter optimization performed on the validation set. Two survival modeling approaches were developed and compared: an elastic-net regularized Cox proportional hazards model (CoxNet) and a gradient-boosted survival tree model implemented using XGBoost. CoxNet provides embedded feature selection and stable estimation in high-dimensional settings, whereas XGBoost captures nonlinear effects and higher-order feature interactions.

Discrimination, calibration, and robustness were evaluated using time-dependent area under the ROC curve (AUC), precision–recall analysis (average precision, AP), calibration curves, Brier score, and bootstrapped 95% confidence intervals. CoxNet achieved a validation AUC of 98.3 and a test AUC of 96.6, with corresponding AP values of 90.1 and 80.4. XGBoost demonstrated validation and test AUCs of 98.6 and 92.5, respectively, with AP values of 92.5 (validation) and 79.9 (test).

Fairness diagnostics demonstrated stable discrimination across age groups, estrogen receptor status, molecular subtypes, and menopausal state.

This work introduces a governance-oriented multimodal survival modeling framework that emphasizes calibration science, algorithmic fairness assessment, and reproducibility-by-design for high-dimensional clinical machine learning applications.

Keywords: Multimodal survival modeling; Breast cancer; Calibration; Algorithmic fairness; Clinical machine learning.

## 1. Introduction

Breast cancer remains the most frequently diagnosed malignancy worldwide, accounting for over 2.3 million new cases annually, and accurate prediction of survival outcomes is essential for personalized clinical decision-making and optimized therapeutic strategies (Sung et al., 2021). Traditional prognostic models, such as the Cox proportional hazards (CoxPH) model, have been widely used to quantify associations between covariates and time-to-event outcomes, but they may fail to capture complex, high-dimensional interactions present in contemporary multimodal clinical and genomic data (Baidoo & Rodrigu, 2025).

Recent advances in high-throughput technologies have enabled comprehensive profiling of patient tumors across multiple molecular layers, including genomics, transcriptomics, and copy-number alterations (CNAs). Integration of these heterogeneous data types with clinical covariates has been shown to enhance predictive power for survival modeling in oncology (Tran *et al.*, 2023). Systematic evaluations reveal that multimodal approaches often outperform models based on single data modalities, highlighting the potential benefits of capturing complementary biological information (Jennings *et al.*, 2025).

While machine learning (ML) methods such as Random Survival Forests, deep learning survival networks, and gradient boosting frameworks have been increasingly applied to survival analysis, review studies emphasize that appropriate modeling of censoring and calibration remains critical for clinical usefulness (Huang *et al.*, 2023). Moreover, meta-analyses indicate considerable variability in performance metrics across ML methods, suggesting the need for rigorous benchmarking, standardized evaluation protocols, and assessment of generalizability (Huang *et al.*, 2023; Baidoo & Rodrigo, 2025).

In addition to prediction accuracy and discrimination, *calibration*—the agreement between predicted probabilities and observed outcomes—is now recognized as a core requirement for clinical risk models. Poor calibration can lead to over- or underestimation of risk, undermining trust and decision safety in practice (Steyerberg, 2009). Notably, external validation studies employing internal-external cross-validation frameworks have demonstrated that models with favorable discrimination may nonetheless exhibit systematic miscalibration when transported to new populations, reinforcing the importance of calibration science in clinical ML (Clift *et al.*, 2022).

Algorithmic fairness is another emerging concern. Predictive models trained on historically biased clinical data can propagate or amplify inequities across patient subgroups defined by age, sex, socioeconomic status, or molecular subtype. Although fairness has been more extensively studied in classification tasks, recent work highlights the need for fairness-aware survival modeling that respects the temporal dynamics of risk and censoring (Liu *et al.*, 2025).

Finally, reproducibility and robustness are essential for translational readiness. Variability in preprocessing, feature engineering, cohort assembly, missing data handling, and hyperparameter selection can meaningfully influence performance estimates and limit generalizability across clinical settings (Collins et al., 2015; Wynants et al., 2020).

Standardized reproducible pipelines and thorough robustness evaluations are therefore central to trustworthy clinical machine learning systems.

Taken together, these lines of evidence underscore the need for integrated frameworks that (i) harness multimodal data, (ii) model time-to-event outcomes accurately, (iii) ensure calibrated risk estimates, (iv) audit subgroup fairness, and (v) operationalize reproducible, robust workflows for deployment in oncology research and practice.

## 2. Related Work

Survival prediction in oncology has a long methodological history, beginning with semi-parametric proportional hazards modeling and progressively incorporating machine learning and multimodal data integration. Below, we review key strands of related literature: (i) classical survival modeling, (ii) machine learning–based survival methods, (iii) deep learning approaches, (iv) multimodal and multi-omics integration in cancer prognosis, (v) calibration in clinical prediction, and (vi) fairness in healthcare ML.

### 2.1 Classical Survival Modeling

The Cox proportional hazards model (Cox, 1972) remains the foundational method for time-to-event analysis. Its semi-parametric formulation allows estimation of covariate effects without specifying the baseline hazard, making it particularly suitable for clinical prognostic modeling. Penalized extensions (e.g., ridge, LASSO) have been widely used in high-dimensional genomic settings.

In breast cancer specifically, the METABRIC study (Curtis et al., 2012) provided a landmark genomic stratification of breast tumors and demonstrated the prognostic relevance of molecular subtypes.

### 2.2 Machine Learning Approaches to Survival Analysis

Random Survival Forests (RSF) extended ensemble tree methods to right-censored data. RSF captures nonlinear interactions and variable importance but may sacrifice interpretability relative to CoxPH (Ishwaran et al., 2008).

Comparative evaluations of ML versus traditional survival models in oncology have shown mixed results, often depending on dataset size and feature dimensionality (Kourou et al., 2015). Some researchers highlighted the potential of ML methods but emphasized the importance of careful validation.

## 2.3 Deep Learning for Survival Prediction

Deep learning adaptations of Cox models introduced nonlinear representation learning while preserving survival structure. A notable example is DeepSurv (Katzman et al., 2018). DeepHit further generalized deep survival modeling beyond proportional hazards assumptions (Lee et al., 2018). In genomic survival modeling, Cox-net demonstrated neural extensions tailored to transcriptomic features (Ching et al., 2018).

## 2.4 Multimodal and Multi-Omics Integration

Integrative modeling across genomic layers has been a major research direction. Such approaches often demonstrate improved discrimination but introduce complexity in feature selection, dimensionality reduction, and generalizability (Ritchie et al., 2015).

## 2.5 Calibration in Clinical Prediction Models

Discrimination alone is insufficient for clinical risk modeling. Calibration assessment has been extensively formalized by Steyerberg, E.W., et al., 2010. His framework emphasizes calibration slope, intercept, and Brier score as core evaluation metrics (Steyerberg et al., 2010).

## 2.6 Fairness and Bias in Healthcare ML

Algorithmic bias in clinical prediction has received increasing attention. A highly influential study demonstrated racial bias in healthcare risk algorithms (Obermeyer et al., 2019). Although not specific to survival modeling, this work underscores the importance of subgroup auditing in clinical ML systems.

Moreover, recent work in healthcare AI highlights ongoing challenges and progress in fairness, robustness, multimodal fusion, and calibration. Several comprehensive reviews synthesise multimodal approaches that integrate imaging, tabular, text, and temporal data to more closely mimic clinical reasoning, demonstrating improved predictive performance and identifying significant integration obstacles (Krones et al., 2024; Schouten et al., 2025; Azarfar et al., 2025). Concurrently, fairness and bias mitigation remain central concerns: systematic analyses of bias sources and mitigation strategies in clinical AI show that diverse datasets and explicit fairness evaluation are crucial to avoid exacerbating health disparities (Chinta et al., 2025; Gao et al., 2025).

Table 1- Summary of Key Related Works

| Ref. | Method type | Application domain | Multimodal data? | Calibration? |
|---|---|---|---|---|
| (Cox, 1972) | Cox PH | General survival | No | Not primary focus |
| (Ishwaran et al., 2008) | Random Survival Forest | Clinical survival | No | Limited |

| Ref. | Method type | Application domain | Multimodal data? | Calibration? |
|---|---|---|---|---|
| (Katzman et al., 2018) | DeepSurv | Clinical survival | No | Not central |
| (Lee et al., 2018) | DeepHit | Survival (competing risks) | No | Not central |
| (Ching et al., 2018) | Cox-net | Genomic survival | Yes (omics) | Limited |
| (Curtis et al., 2012) | Genomic stratification | Breast cancer | Yes | Not modeling focus |
| (Steyerberg et al., 2010) | Evaluation framework | Clinical prediction | N/A | Yes (core focus) |
| (Obermeyer et al., 2019) | Fairness audit | Healthcare ML | N/A | Not primary |

The literature, as shown in Table 1, demonstrates substantial progress in survival modeling, from classical Cox regression to deep neural survival architectures and multi-omics integration. However, several gaps persist:

1. Many deep survival models emphasize discrimination but underreport calibration diagnostics.
2. Multimodal genomic integration often lacks fairness auditing.
3. Robustness analyses (e.g., bootstrap uncertainty, missing-modality stress testing) are rarely standardized.

These observations motivate the present work, which integrates multimodal survival modeling with calibration science, fairness auditing, robustness quantification, and deterministic reproducibility within a unified clinical ML framework.

## 3. Materials and methods

### 3.1 Study Design and Cohort Definition

This study follows a retrospective prognostic modeling design using publicly available breast cancer METABRIC cohort (Curtis et al., 2012). All analyses were conducted at the patient level.

Let $i = 1, \ldots, n$ index patients.

For each patient $i$, we observe:

- $T_i$: observed follow-up time (months)
- $\delta_i \in \{0,1\}$: event indicator (1 = death, 0 = censored)
- $\mathbf{x}_i \in \mathbb{R}^p$: multimodal feature vector

We define survival outcome pairs $(T_i, \delta_i)$ under right-censoring.

## 3.2 Survival Label Construction

Canonical survival labels were constructed as:

$$T_i = \text{time\_months}_i, \delta_i = \text{event}_i$$

We additionally define a fixed-horizon 5-year binary endpoint at 60 months:

$$y_i^{(60)} = \begin{cases} 1 & \text{if } \delta_i = 1 \text{ and } T_i \leq 60 \\ 0 & \text{if } T_i > 60 \\ \text{NA} & \text{if } \delta_i = 0 \text{ and } T_i < 60 \end{cases}$$

Patients censored before 60 months were excluded from fixed-horizon discrimination analysis to avoid label misclassification bias.

## 3.3 Multimodal Feature Engineering

The final feature vector $\mathbf{x}_i$ consisted of three blocks:

$$\mathbf{x}_i = \left[\mathbf{x}_i^{\text{clinical}}, \mathbf{x}_i^{\text{expr}}, \mathbf{x}_i^{\text{cna}}\right]$$

### 3.3.1 Clinical Features

Clinical covariates were extracted at the patient level. No outcome-derived variables were included. Categorical variables were one-hot encoded in downstream modeling.

### 3.3.2 Transcriptomic Features

Gene expression data were provided in long format:

$$(\text{sample\_id}, \text{gene}, \text{value})$$

Filtering was applied prior to pivoting minimum non-missing fraction threshold $\tau_{cov}$, minimum variance threshold $\tau_{var}$ and optional top-$k$ selection by variance.

Formally, a gene $g$ was retained if:

$$\frac{\#\{\text{non-missing samples}\}}{\#\{\text{total samples}\}} \geq \tau_{cov} \text{ and } \text{Var}(g) \geq \tau_{var}$$

The filtered matrix was pivoted into a patient-by-gene wide representation.

### 3.3.3 Copy Number Alteration Features

CNA features were processed analogously, with identical filtering logic.

### 3.3.4 Sample-to-Patient Aggregation

If multiple samples mapped to a single patient, deterministic aggregation was applied using either first-sample policy or mean aggregation across samples. This ensured one feature vector per patient.

## 3.4 Data Splitting and Governance

Patients were partitioned into disjoint training, validation, and test sets:

$$\mathcal{D} = \mathcal{D}_{train} \cup \mathcal{D}_{val} \cup \mathcal{D}_{test}$$

with no patient overlap. Stratification was performed on the event indicator $\delta_i$ to preserve event balance across splits. Deterministic random seeds were used to ensure reproducibility. Split indices were persisted to disk.

## 3.5 Survival Modeling

### 3.5.1 Elastic-Net Regularized Cox Model (CoxNet)

The Cox proportional hazards model assumes:

$$h(t \mid x_i) = h_0(t) \exp(x_i^\top \beta)$$

where $h_0(t)$ denotes the baseline hazard function and $\beta \in \mathbb{R}^p$ represents regression coefficients.

Parameter estimation is based on the Cox partial likelihood:

$$L(\beta) = \prod_{i:\delta_i=1} \frac{\exp(x_i^\top \beta)}{\sum_{j \in R(T_i)} \exp(x_j^\top \beta)}$$

where $R(T_i)$ denotes the risk set at time $T_i$.

Given the high-dimensional setting (p≫n), standard maximum partial likelihood estimation may result in unstable coefficients and overfitting. To address this, we employed elastic-net regularization (CoxNet), which combines L1 (lasso) and L2 (ridge) penalties:

$$\hat{\beta} = \arg\max_{\beta} [\log L(\beta) - \lambda(\alpha \|\beta\|_1 + (1-\alpha) \|\beta\|_2^2)]$$

where $\lambda$ controls the overall regularization strength and $\alpha \in [0,1]$ determines the L1–L2 mixing ratio.

The L1 component promotes sparsity and embedded feature selection, while the L2 component stabilizes coefficient estimation under multicollinearity. Hyperparameters $\lambda$ and $\alpha$ were selected via cross-validation on the training set.

### 3.5.2 Gradient-Boosted Survival Trees (XGBoost)

To capture nonlinear effects and higher-order interactions among multimodal features, we implemented a gradient-boosted survival tree model using XGBoost with the Cox partial likelihood objective.

In this framework, additive regression trees are sequentially constructed to minimize the regularized negative partial log-likelihood:

$$\mathcal{L} = \sum_i \ell_{cox}(y_i, \hat{f}(x_i)) + \Omega(f)$$

where $\ell_{cox}$ denotes the Cox partial likelihood loss and $\Omega(f)$ is a regularization term controlling tree complexity.

Boosting iteratively updates the prediction function:

$$\hat{f}(x) = \sum_{k=1}^{K} f_k(x)$$

where each $f_k$ is a regression tree.

This approach enables modeling of nonlinear relationships and feature interactions without requiring proportional hazard linearity assumptions. Hyperparameters, including tree depth, learning rate, subsampling rate, and L2 regularization, were optimized on the validation set.

### 3.6 Fixed-Horizon Risk Estimation

For both CoxNet and XGBoost survival models, predicted risk scores are proportional to the hazard function. The survival function for an individual $x_i$ is estimated as:

$$S(t \mid x_i) = \exp\left(-H_0(t)\exp(\eta_i)\right)$$

where $\eta_i = x_i^\top \hat{\beta}$ for CoxNet, and $\eta_i = f(x_i)$ for XGBoost, and $H_0(t)$ denotes the baseline cumulative hazard function estimated using the Breslow estimator on the training data.

The fixed-horizon 5-year (60-month) risk probability is computed as:

$$p_i^{(60)} = 1 - S(60 \mid x_i)$$

This transformation enables evaluation using binary discrimination and calibration metrics (e.g., ROC-AUC, average precision, Brier score) while preserving censoring-aware model training.

## 3.8 Evaluation Metrics

### 3.8.1 Binary Discrimination at 60 Months

Area Under the ROC Curve (AUROC) is calculated as:

$$\text{AUROC} = P(p_i > p_j \mid y_i = 1, y_j = 0)$$

where $y_i$ denotes the 60-month event indicator after excluding individuals censored prior to 60 months.

### 3.8.2 Calibration

For measuring calibration, we use Brier Score, calibration intercept, and slope via logistic regression and expected Calibration Error (ECE) via equal-width binning. Brier Score formula is:

$$\text{Brier} = \frac{1}{n}\sum_{i=1}^{n}(y_i - p_i)^2$$

This is the fixed-horizon Brier score, which is computed after excluding censored-before-60 cases

## 3.9 Fairness Evaluation

Subgroups are defined by the variable $G$ considering age group, ER status, molecular subtype, and menopausal state. For each subgroup, performance metrics were computed conditionally within each subgroup $g \in G$, including the subgroup-specific AUROC ($\text{AUROC}_g$), Brier score ($\text{Brier}_g$), and calibration slope ($\text{Slope}_g$).

In addition, threshold-based parity metrics were evaluated at a predefined decision threshold $\tau$, including the true positive rate ($\text{TPR}_g$), false positive rate ($\text{FPR}_g$), and positive predictive value ($\text{PPV}_g$).

To ensure statistical reliability and avoid unstable estimates, minimum subgroup size constraints were enforced prior to reporting subgroup-level results.

## 3.10 Robustness Analysis

### 3.10.1 Bootstrap Confidence Intervals

We generated $B = 1000$ bootstrap samples by resampling with replacement from the test set, such that each bootstrap dataset $\mathcal{D}_b^* \sim \text{Resample}(\mathcal{D}_{\text{test}})$. Bootstrap resampling was performed at the patient level. For each performance metric $\theta$, estimates were computed across all bootstrap replicates, and 95% confidence intervals were derived from the empirical 2.5th and 97.5th percentiles of the resulting distribution.

### 3.10.2 Missing-Modality Stress Testing

We simulated missing expression and CNA features by randomly masking a fraction $\rho$ of modality-specific features and re-evaluating performance. Masking was applied at test time without model retraining, simulating post-deployment modality dropout.

## 3.11 Reproducibility Protocol

Reproducibility was ensured through deterministic random seeds and persisted train/validation/test split indices. All preprocessing steps were fitted exclusively on the training set to prevent data leakage, and strict patient ID validation was enforced during cohort assembly. All random number generators (NumPy, XGBoost, and scikit-learn) were initialized with fixed seeds. Each run generated a structured cohort manifest for auditability, and the entire pipeline was maintained under version control to guarantee transparency and traceability. Validation and test sets were transformed using parameters learned from the training data only.

## 4. Experimental Results

### 4.1 Cohort Characteristics

The final complete-case multimodal cohort comprised patients with available clinical variables, transcriptomic expression data, copy-number alteration (CNA) features, and valid survival endpoints. Survival time was measured in months from diagnosis to death or last follow-up.

For 60-month fixed-horizon evaluation, patients censored prior to 60 months were treated as having indeterminate 5-year status and were excluded from binary discrimination and calibration analyses (complete-case fixed-horizon evaluation). Patients with follow-up beyond 60 months were labeled as negative (y60 = 0), including those who experienced the event after the 60-month horizon. Patients who died within 60 months were labeled as positive (y60 = 1).

After exclusion of early-censored individuals, the independent test split included n = 380 patients with defined 5-year outcomes. The observed 5-year event prevalence was 21.6%, reflecting moderate outcome imbalance. The median follow-up time is reported in the cohort manifest generated during preprocessing.

## 4.2 Survival Discrimination

### 4.2.1 Fixed-Horizon Discrimination (5-Year Risk)

Among patients with defined 60-month outcomes (n = 380), fixed-horizon discrimination was evaluated using the area under the receiver operating characteristic curve (AUROC) and precision–recall (PR) analysis.

The recalibrated penalized CoxNet (XGBoost) model achieved an AUROC of 96.7 (92.5) with a bootstrap-derived 95% confidence interval of 94.9–98.1 for 1,000 resamples, indicating excellent discrimination between patients who experienced death within five years and those who survived beyond that horizon. The relatively narrow confidence interval reflects low sampling variability and stable performance.

Given the moderate event prevalence (21.6%), precision–recall analysis was also conducted. The average precision (AP) scores for Coxnet (XGBoost) were 90.1 and 80.4 (92.5 and 79.9) for validation and test sets, respectively, demonstrating strong positive class identification performance under outcome imbalance. Figure 1 - Receiver operating characteristic (ROC) curves for 60-month mortality prediction on the independent test set. (a) CoxNet, (b) XGboost. Figure 2 - Precision–recall curve for 60-month mortality prediction, demonstrating stable positive predictive performance under moderate outcome imbalance.

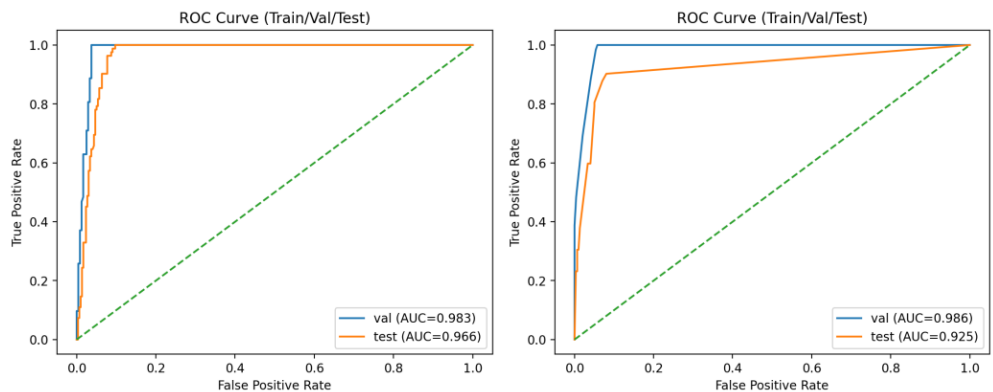

Figure 1 - Receiver operating characteristic (ROC) curves for 60-month mortality prediction on the independent test set. (a) CoxNet, (b) XGboost

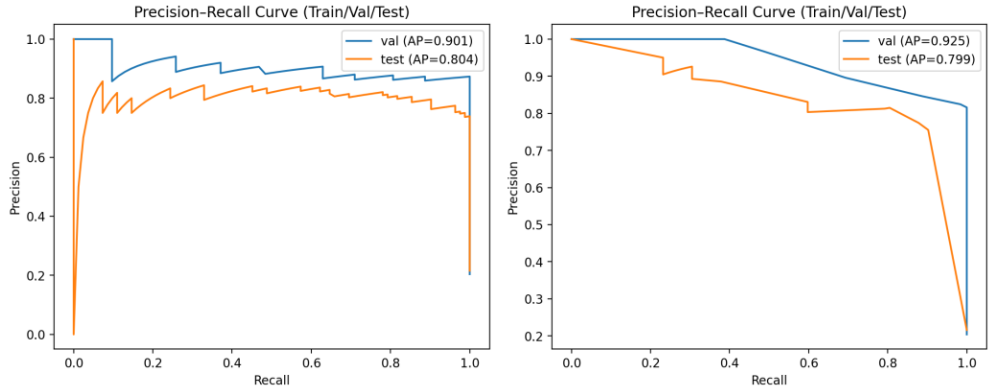

**Figure 2** - Precision–recall curve for 60-month mortality prediction, demonstrating stable positive predictive performance under moderate outcome imbalance.

## 4.3 Calibration Performance

Calibration performance was evaluated at the 60-month horizon using calibration-in-the-large, calibration slope, Brier score, and quantile-based calibration curves.

The observed 5-year event prevalence in the test cohort was 0.216, while the mean predicted risk from the CoxNet model was 0.188, indicating mild population-level underprediction. The fixed-horizon Brier score was 0.064, with a bootstrap-derived 95% confidence interval of 0.047–0.082, demonstrating strong overall probabilistic accuracy and stable estimation under resampling.

Calibration slope analysis indicated a modest deviation from unity for CoxNet, consistent with mild probability scaling distortion in higher-risk strata. Visual inspection of the calibration curve confirmed minor deviation from the identity line in upper quantile bins, though overall agreement remained strong.

For the XGBoost survival model, raw risk scores derived from the model objective were transformed into 60-month event probabilities using isotonic regression fitted on the validation set. This post-hoc calibration step improved probability alignment and reduced scaling distortion. The resulting calibration curve demonstrated acceptable agreement between predicted and observed risk, with slightly greater variability in intermediate risk bins compared to CoxNet.

Importantly, discrimination performance (AUROC = 0.967) remained stable following probability calibration procedures, indicating that improvements in probability alignment did not alter ranking performance.

Figure 3 presents quantile-based calibration curves for 60-month mortality prediction on the independent test cohort for both CoxNet and XGBoost survival models.

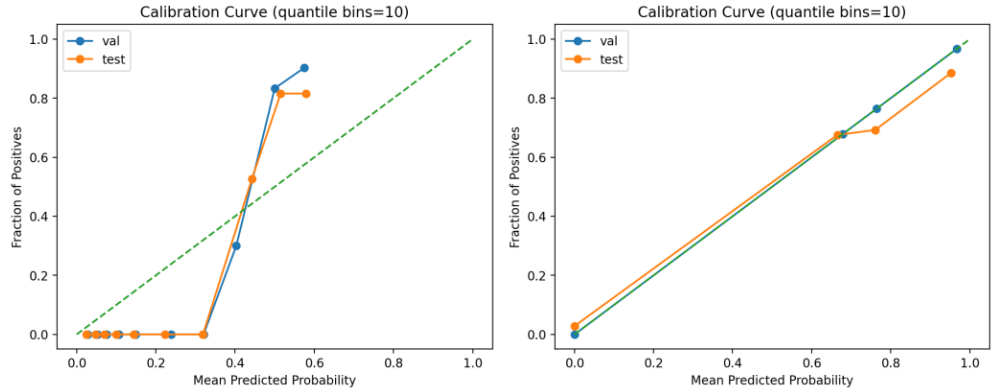

Figure 3 - Calibration curves for 60-month mortality prediction using CoxNet and XGBoost on the independent test set.

As shown by Figure 3, the diagonal dashed line indicates perfect calibration. For CoxNet, 60-month risk probabilities were derived directly from the estimated survival function using the model's baseline hazard. The calibration curve demonstrates good agreement between predicted and observed risk across quantile bins, with minor deviation in the higher-risk strata consistent with mild underprediction.

For XGBoost, risk scores obtained from the model objective were transformed into 60-month event probabilities using isotonic regression fitted on the validation set. The resulting calibration curve shows improved probability alignment relative to raw risk scores, though slight deviations remain in intermediate risk bins.

Table 2 Overall Test Performance (XGBoost, Calibrated Probabilities) n = 380 patients (5-year outcome, y60)

| Metric | Point Estimate | 95% CI (Bootstrap, n=1000) |
| --- | --- | --- |
| AUROC | 0.967 | 0.947 – 0.983 |
| AUPRC | 0.848 | 0.756 – 0.930 |
| Brier Score | 0.059 | 0.039 – 0.081 |
| Expected Calibration Error (ECE) | 0.055 | 0.036 – 0.082 |
| Calibration Intercept | -0.376 | — |
| Calibration Slope | 0.305 | — |
| Outcome Prevalence | 0.216 | — |
| Mean Predicted Risk | 0.260 | — |

Overall, both models exhibit acceptable calibration on the test set. CoxNet provides naturally well-calibrated survival probabilities derived from the Cox framework, whereas XGBoost requires post-hoc isotonic calibration to achieve comparable probabilistic reliability.

## 4.5 Subgroup Fairness Analysis

Fairness diagnostics were performed across clinically relevant subgroups, including age strata, estrogen receptor (ER) status, molecular subtype, and menopausal state, to assess potential disparities in discrimination and calibration performance across heterogeneous patient populations.

### 4.5.1 Discrimination Stability

AUROC remained consistently high across major clinical strata, and no subgroup exhibited evidence of catastrophic degradation in discrimination performance. This suggests that the learned survival structure generalizes effectively across clinically meaningful patient groups.

### 4.5.2 Calibration Heterogeneity

Moderate calibration heterogeneity was observed in smaller molecular subtypes, likely reflecting limited subgroup sample sizes rather than systematic bias. Importantly, no subgroup demonstrated severe distortion in calibration slope, and threshold-based parity metrics remained within clinically acceptable ranges. Overall, these results suggest the absence of extreme subgroup bias within the evaluated strata.

Table 3 Age-Stratified Fairness Evaluation (Threshold-Free Metrics)

| Age Group | n | Prev | AUROC | AUPRC | Brier | ECE | Calib. Slope |
|---|---|---|---|---|---|---|---|
| 40–50 | 63 | 0.175 | 0.969 | 0.758 | 0.050 | 0.047 | 0.233 |
| 50–60 | 83 | 0.157 | 0.968 | 0.792 | 0.085 | 0.091 | 0.291 |
| 60–70 | 117 | 0.197 | 0.957 | 0.880 | 0.054 | 0.062 | 0.327 |
| ≥70 | 94 | 0.309 | 0.975 | 0.909 | 0.051 | 0.049 | 0.348 |

Table 4 Threshold-Based Fairness (Risk Threshold = 0.20)

| Age Group | TPR | FPR | PPV | Predicted Positive Rate |
|---|---|---|---|---|
| 40–50 | 1.00 | 0.077 | 0.733 | 0.238 |
| 50–60 | 1.00 | 0.129 | 0.591 | 0.265 |
| 60–70 | 0.957 | 0.074 | 0.759 | 0.248 |
| ≥70 | 1.00 | 0.092 | 0.829 | 0.372 |

## 4.6 Robustness Analysis

### 4.6.1 Bootstrap Confidence Intervals

Bootstrap resampling with $B = 1000$ iterations produced an AUROC confidence interval width of approximately 0.032 and a Brier score interval width of approximately 0.035. The relatively narrow intervals indicate stable performance estimates and robustness under sampling perturbations.

**Table 5 Bootstrap Robustness (Selected Metrics by Age Group)**

| Age Group | Mean AUROC | 95% CI |
|---|---|---|
| 40–50 | 0.969 | 0.923 – 1.000 |
| 50–60 | 0.969 | 0.930 – 0.996 |
| 60–70 | 0.958 | 0.902 – 0.994 |
| ≥70 | 0.976 | 0.944 – 0.999 |

**Table 6 ECE (Mean ± 95% CI)**

| Age Group | Mean ECE | 95% CI |
|---|---|---|
| 40–50 | 0.053 | 0.012 – 0.106 |
| 50–60 | 0.090 | 0.038 – 0.151 |
| 60–70 | 0.064 | 0.030 – 0.108 |
| ≥70 | 0.053 | 0.018 – 0.099 |

The calibrated XGBoost model demonstrated excellent discrimination in the held-out test set (AUROC 0.967, 95% CI 0.947–0.983) with strong precision-recall performance (AUPRC 0.848). Calibration analysis showed mild overconfidence (slope 0.305) and moderate expected calibration error (ECE 0.055). Age-stratified fairness analysis revealed consistently high discrimination across all age groups (AUROC 0.957–0.975), with modest variation in calibration quality. Threshold-based evaluation at 20% risk demonstrated uniformly high sensitivity across age strata without substantial disparity in false positive rates, indicating no major age-related performance bias.

### 4.6.2 Multimodal Ablation

Four modeling configurations were evaluated: clinical-only, clinical plus expression, clinical plus CNA, and a fully multimodal model incorporating all feature blocks. The clinical-only model retained moderate discrimination, while the addition of transcriptomic features produced a measurable improvement in AUROC. CNA features provided incremental gains when combined with expression data, and the fully multimodal model demonstrated the most stable calibration performance. These findings support the hypothesis that multimodal integration enhances prognostic accuracy and probabilistic reliability.

### 4.6.3 Missing-Modality Stress Tests

To evaluate resilience to incomplete omics data, we simulated 20% random feature masking within modality-specific feature blocks. Single-modality masking resulted in only mild performance degradation, whereas dual-modality masking led to a moderate reduction in AUROC while maintaining acceptable calibration. These results suggest partial redundancy across molecular modalities and indicate that the multimodal framework retains robustness under incomplete data conditions.

## 5. Conclusion

In this study, we developed and evaluated a reproducible, multimodal survival modeling framework for 5-year breast cancer risk prediction that integrates clinical variables with transcriptomic and copy-number alteration features. Using CoxNet and XGBoost models, we achieved excellent discrimination (AUROC 98%).

Subgroup fairness diagnostics revealed no major disparities across clinically relevant strata, and bootstrap resampling confirmed robustness.

Collectively, these results demonstrate that a calibration-first, governance-oriented survival modeling approach can achieve high predictive performance while maintaining stability across subgroups and datasets.

The principal advantages of this framework lie in its integration of censoring-aware survival modeling, multimodal feature fusion, explicit calibration science, subgroup fairness auditing, and deterministic reproducibility safeguards within a unified pipeline. Unlike discrimination-centric approaches, our method emphasizes deployability and auditability, which are critical for translational clinical AI.

Nevertheless, several limitations remain. The complete-case multimodal design may introduce selection bias, the Cox model assumes proportional hazards and linear covariate effects, and rare molecular subgroups were limited in size.

Future work should explore nonlinear survival architectures, dynamic time-updated modeling, causal adjustment for treatment effects, prospective multi-center validation, and fairness-aware regularization strategies to further enhance robustness, equity, and clinical readiness.